\documentclass[11pt]{article}
\usepackage[T1]{fontenc}
\usepackage[utf8]{inputenc}
\DeclareUnicodeCharacter{02BF}{\textquotesingle}
\DeclareUnicodeCharacter{1EB7}{\d{\u{a}}}
\usepackage{times}
\usepackage{latexsym}
\usepackage{microtype}
\usepackage{inconsolata}
\usepackage{graphicx}
\usepackage{booktabs}
\usepackage{multirow}
\usepackage{amsmath}
\usepackage{enumitem}
\usepackage{xspace}
\usepackage[preprint]{acl}
\title{A Comparative Study: Which Models Perform Better in Inheritance Reasoning?}

\author{
  Mohammed Lamine Mouhoub \\
  Paris Dauphine University \\
  \texttt{\small{mohamed.mouhoub@dauphine.fr}}
  \And
  Chahinez Bouchekif \\
  University of Abou Bekr Belkaïd \\
  \texttt{\small{meriemchahinez.bouchekif@univ-tlemcen.dz}}
}

\begin{document}
\maketitle
\begin{abstract}
This paper presents the participation of team PSL in the QIAS 2026 Shared Task on Arabic Islamic inheritance reasoning. The task evaluates the ability of large language models to solve inheritance cases that require legal interpretation, multi-step reasoning, and precise numerical computation. We compare \textit{commercial} and \textit{open-source} models under a unified prompting strategy to assess their effectiveness in structured legal reasoning with minimal task-specific adaptation.

Our results show a clear gap in reliability between the two model families. Commercial models demonstrate stronger performance in identifying eligible heirs, applying exclusion rules, and maintaining consistency across reasoning steps. In contrast, open-source models exhibit greater instability, particularly in cases involving dependent legal decisions and fractional share adjustments. The best performance is achieved by \textit{Gemini 2.5 Flash}, with an MRE of $0.989$.

\end{abstract}
\section{Introduction}

Large language models (LLMs) perform well on many NLP tasks \cite{hassan2023firc,jahan2023multilingual}, but this does not always mean that they can reason well in rule-based domains. In such tasks, a good answer must be both plausible and correct at every step. If the model makes one wrong decision early in the process, the whole solution may become incorrect. For this reason, more attention is now being given to evaluation tasks that test reasoning with structured rules, symbolic relations, and numerical accuracy.

Islamic inheritance law (\textit{ʿilm al-mawārīth}) is a particularly challenging example of this problem. Solving an inheritance case requires more than retrieving legal knowledge or recognizing common patterns. A correct solution must identify the eligible heirs, apply blocking and exclusion rules correctly, assign the prescribed shares, and decide whether adjustment mechanisms such as \textit{ʿawl} or \textit{radd} are needed. Because all of these decisions are closely connected, inheritance reasoning is a useful benchmark for studying whether current LLMs can perform coherent multi-step legal reasoning rather than simply generate fluent legal language \citep{bouchekif2026mawarith}.

Recent work on reasoning-oriented large language models has highlighted the importance of evaluating models on tasks that require more than fluent text generation. Chain-of-thought prompting improves performance on multi-step problems by encouraging models to produce intermediate reasoning steps, while self-consistency increases robustness by selecting answers that are consistent across multiple sampled reasoning paths~\citep{wei2022chain, wang2022self}. More recent reasoning-focused systems, such as OpenAI's o1 family and DeepSeek-R1, further suggest that additional inference-time reasoning or reinforcement learning can lead to strong gains on difficult reasoning tasks~\citep{openai2024o1, guo2025deepseek}. However, these advances do not fully solve the problem. Reasoning remains fragile in tasks that require strict rule application, long chains of dependent decisions, or exact numerical computation, where a single intermediate mistake may invalidate the final answer~\citep{wei2022chain, wang2022self}.

This limitation is especially clear in specialized legal tasks such as Islamic inheritance, where correct answers depend on both legal reasoning and precise numerical calculation. In such settings, reasoning-oriented models often perform better than standard LLMs. Several studies on the QIAS 2025 benchmark \cite{bouchekif2025qias} showed that models such as Gemini and ChatGPT consistently outperformed many Arabic-focused and open-source models, including Fanar and ALLaM, on inheritance-law evaluation~\citep{bouchekif2025assessing, aldahoul-zaki-2025-nyuad-qias, al-smadi-2025-qu, bekhouche2025cvpd}. Notable exceptions include \citet{elrefai-etal-2025-gumball}, who obtained the best results using Qwen3, and \citet{xuan-phuc-dang-van-2025-puxai}, who proposed a hybrid multi-agent architecture.

In this paper, we describe the participation of team PSL in QIAS 2026. Our submission compares commercial and open-source models under a unified prompting framework. Instead of designing a highly specialized task-specific pipeline, we examine a simpler and more informative question: how well can current general-purpose models solve a structured Arabic legal reasoning task when evaluated under the same conditions. This comparison is important because it helps clarify the gap between highly capable proprietary systems and more accessible open-weight models.

\section{Data}

Our experiments are conducted on the QIAS 2026 benchmark \cite{bouchekif2026qias}. The dataset contains 12,500 inheritance cases written in natural language and follows the majority juristic opinion (\textit{al-jumhūr}). Each instance describes the deceased and the surviving relatives, and the task is to determine the eligible heirs and their final shares according to the rules of Islamic inheritance. Each case provides all the information required to solve the problem, including the inheriting heirs, blocked heirs, assigned shares, the possible application of \textit{ʿawl} or \textit{radd}, and the final normalized distribution. In addition, the dataset includes an intermediate reasoning trace and a concise final answer, making it suitable for evaluating both reasoning quality and final prediction accuracy.
The full corpus is divided into 12,000 training instances and 500 test instances. It covers 36 distinct heir categories, ranging from close relatives such as parents, children, and spouses to more distant agnatic relatives across multiple generations. The cases vary in difficulty, from simple configurations with a small number of heir types to more complex scenarios involving up to twelve distinct heir categories. In terms of legal complexity, the training split contains 11,079 simple cases, 577 \textit{ʿawl} cases, and 344 \textit{radd} cases. The totals reported in the paper imply that the 500-case test split contains 456 simple cases, 39 \textit{ʿawl} cases, and 5 \textit{radd} cases. This distribution shows that most examples do not require adjustment, while a smaller but important subset evaluates the model’s ability to handle more difficult redistribution and proportional-reduction cases. 

\section{System Overview}

Our submission uses a simple prompting-based chain-of-thought (CoT) design. The goal is not to build a complex task-specific system, but to compare different types of models in a fair setting. In particular, we compare reasoning and non-reasoning models, as well as commercial and non-commercial models, under the same conditions. This allows us to better assess their actual ability to solve Arabic inheritance reasoning problems.

Given an inheritance case in Arabic, the system prompts the model to produce a structured solution. The prompt encourages the model to proceed through the main inheritance stages: identifying heirs, applying blocking rules, determining legal shares, and producing the final answer. To reduce unnecessary variation in generation, we use a constrained output format and a small amount of post-processing.

\subsection{Models}

We evaluated both commercial and open-source models to better understand the current gap between these two groups on a structured Arabic legal reasoning task. The models were selected for three main reasons. First, we wanted a fair comparison between systems that are widely accessible to researchers. Second, we wanted diversity in model families and capabilities. Third, we wanted to test whether strong general reasoning models can transfer well to a domain that combines Arabic input, legal constraints, and precise numerical output.

\subsubsection{Commercial Models}

Our commercial systems included \textit{Gemini 2.5 Flash} and \textit{Gemini 2.5 Pro}. These models were selected for their strong general-purpose reasoning capabilities and solid performance in applied NLP tasks. They provide a realistic baseline for systems that users may rely on when addressing legal or knowledge-intensive problems without task-specific fine-tuning.

\subsubsection{Open-Source Models}

Our open-source systems included \textit{Qwen3-32B}\footnote{\href{https://huggingface.co/Qwen/Qwen3-32B}{Qwen3-32B model page}}, \textit{GPT-oss-120B}\footnote{\href{https://huggingface.co/openai/gpt-oss-120b}{GPT-oss-120B model page}}, \textit{Llama-3.3-70B}\footnote{\href{https://huggingface.co/meta-llama/Meta-Llama-3-70B}{Llama-3.3-70B model page}}, \textit{Fanar-Sadiq}, and \textit{Fanar-C-2-27B}\footnote{\href{https://api.fanar.qa/docs}{Fanar API documentation}}. These models were chosen to cover different profiles, such as multilingual instruction models, Arabic-oriented systems, and large open-weight reasoning models. We also conducted preliminary experiments with \textit{Falcon}\footnote{\href{https://huggingface.co/tiiuae/Falcon3-10B-Instruct}{Falcon3-10B-Instruct model page}}; however, its outputs frequently failed to follow the prompt instructions, resulting in very poor performance. Consequently, it was excluded from the final evaluation.

\subsection{Prompting Strategy}

Our method is based on a unified prompting framework. Each model receives the inheritance case in Arabic together with instructions asking it to solve the case in a structured way. The prompt explicitly encourages the model to reason step by step and to return an answer that follows a predefined format.

In practice, the prompt asks the model to:
\begin{enumerate}[leftmargin=*,itemsep=1pt]
    \item identify the heirs who inherit,
    \item apply relevant exclusion or blocking rules,
    \item assign the correct legal shares,
    \item and produce the final distribution.
\end{enumerate}

We intentionally kept the prompt relatively simple. The goal was not to encode a handcrafted expert system inside the prompt, but rather to observe how much structured reasoning the models can perform under shared instructions.

For all models, we required the output to show the reasoning at each step, followed by a final answer in a structured format that matches the evaluation metric. This made the outputs more consistent, easier to process, and helped us see exactly where any reasoning mistakes occurred. We only applied simple post-processing to ensure that each heir was represented in a single, uniform form. We did not use extensive correction rules, since our goal was to evaluate the models’ abilities rather than to engineer a complex pipeline.

\section{Results and Analysis}

\begin{table}[t]
\centering
\small
\begin{tabular}{lcc}
\toprule
\textbf{Model} & \textbf{Type} & \textbf{Score} \\
\midrule
Gemini-2.5-Pro & Commercial & 0.931 \\
Gemini-2.5-Flash 2$^*$ & Commercial & 0.898 \\
Qwen3-32B & Open-source & 45.1 \\
GPT-OSS-120B & Open-source & 38.7 \\
LLaMA-3.3-70B & Open-source & 35.9 \\
Fanar-Sadiq & Open-source & 35.8 \\
Fanar-C-2-27B & Open-source & 33.1 \\
\bottomrule
\end{tabular}
\caption{Main results for the PSL systems on QIAS 2026. $^*$ indicates the official submission.}
\label{tab:main_results}
\end{table}
Table~\ref{tab:main_results} reports the overall MIR-E scores obtained by all evaluated models. 
Overall, the results reveal a substantial performance gap between commercial and open-source models (Table~\ref{tab:main_results}). Commercial models achieve significantly higher scores, with \textit{Gemini-2.5-Pro} reaching 0.931 and \textit{Gemini-2.5-Flash 2} (official submission) achieving 0.898. In contrast, open-source models perform considerably worse, with scores ranging from 33.1 to 45.1. The strongest open-source model, \textit{Qwen3-32B}, achieves 45.1, remaining more than 50 points below the official system. Furthermore, performance differences among open-source models are relatively limited, suggesting a broadly similar level of capability within this group.

Beyond aggregate performance, the comparison highlights important qualitative differences in reasoning behavior. Commercial models demonstrate a stronger ability to maintain coherent multi-step reasoning, preserving consistency across heir identification, exclusion rules, and final share allocation. They are also more robust when early-stage decisions are ambiguous or complex. In contrast, open-source models frequently exhibit structural errors at early stages of the reasoning process. Such errors tend to propagate, leading to incorrect intermediate steps and ultimately invalid final distributions. For example, the omission of a valid heir often results in incorrect redistribution of shares.

A further observation is that surface-level fluency does not reliably indicate correctness. Several models produce outputs that appear legally plausible, yet contain latent structural inconsistencies, such as incorrect application of exclusion rules or infeasible share allocations.

Taken together, these results suggest that commercial models are both more accurate and more stable in handling structured legal reasoning tasks, whereas open-source models remain limited in their ability to sustain coherent reasoning across interdependent decision steps.
For comparison, we also consider our method as a participating system in the shared task.

For comparison, we also report the scores of the systems that participated in the shared task.
Table \ref{tab:leaderboard} presents the official ranking of the participating teams. The proposed methods cover a wide range of approaches, from simple prompting-based techniques to fine-tuning strategies that adapt language models to the inheritance task. Some approaches rely on Retrieval-Augmented Generation (RAG) to find similar cases and support reasoning, while others adopt hybrid methods that combine large language models (LLMs) with rule-based systems.
\begin{table}[t]
\centering
\small
\begin{tabular}{cll}
\toprule
\textbf{Rank} & \textbf{Team} & \textbf{MIR-E}  \\
\midrule
1 & CVPD \cite{swaileh2026cvpd} & 0.935  \\
2 & Simplicity \cite{almansour2026simplicity} & 0.931  \\
3 & KMS \cite{alkhamis2026kms}& 0.916 \\
4 & QU-NLP \cite{alsmadi2026qu} & 0.907 \\
5 & PSL (our submission) & 0.898  \\
6 & AGS-KSU \cite{Hicham2026agsksu}& 0.84 \\
7 & Silah \cite{Ghader2026Silah}& 0.826  \\
8 & UTLM & 0.742  \\
\bottomrule
\end{tabular}
\caption{Official leaderboard results for QIAS 2026 for teams that participated in the test phase and submitted a paper}
\label{tab:leaderboard}
\end{table}

\subsection{Common Error Types}

Our manual inspection identified several recurring error patterns:
\paragraph{Missing or hallucinated heirs.}
Some models failed to identify all heirs mentioned in the scenario, while others introduced relatives who were not present in the case. This problem is especially harmful because all later steps depend on the correct heir set.
\paragraph{Incorrect blocking decisions.}
Another common issue was the wrong application of exclusion rules. In some outputs, an heir who should have been blocked was retained; in others, a valid heir was incorrectly removed.

\paragraph{Share assignment errors.}
Even when the correct heirs were identified, some models assigned the wrong legal shares. These errors often reflected confusion between similar family configurations or incomplete understanding of conditional inheritance rules.

\paragraph{Arithmetic inconsistency.}
Some outputs contained legal-sounding reasoning but inconsistent calculations. In such cases, the fractions did not sum correctly, the normalization was missing, or the final per-heir allocation contradicted the earlier explanation.

\paragraph{Format instability.}
A smaller but still relevant issue was output instability. Some models ignored the requested format, mixed reasoning and answer sections unpredictably, or returned partial solutions.

This also explains why good instruction-following is not enough. A model can follow the prompt and produce a clear and convincing explanation, but still make mistakes in the legal structure of the solution.

Another possible reason for the low performance of Arabic-oriented models is that they are not trained on this type of structured reasoning. Inheritance problems require specific logical steps and rule-based decisions, which may not be well represented in their training data. In this context, the QIAS 2026 dataset can be very useful. It provides examples of Islamic inheritance reasoning that could help models learn this type of task and improve their performance.

\section{Conclusion}

This paper presented the participation of team PSL in the QIAS 2026 Shared Task on Arabic Islamic inheritance reasoning. Our submission focused on a controlled comparison between commercial and open-source language models under a shared prompting framework.

The experiments showed that commercial models remain clearly stronger on this task, especially when the case requires several dependent legal decisions and precise numerical consistency. Open-source models show promise, but they still suffer from frequent structural errors that propagate across the reasoning chain. More broadly, our findings confirm that Islamic inheritance is a demanding testbed for evaluating legal reasoning in Arabic.

In future work, we plan to explore stronger output constraints, step-level verification, and domain-adapted training strategies to improve consistency on this class of tasks. We also believe that combining LLMs with explicit reasoning checks may be a promising direction for legal and rule-based NLP more generally.

\section*{Acknowledgments}

We would like to thank the organizers of QIAS 2026 for their efforts in designing and running the shared task. 
\bibliography{lrec2026-example}

\end{document}